\documentclass[runningheads]{llncs}
\usepackage{graphicx}
\usepackage{amsmath,amssymb} % define this before the line numbering.
\usepackage{color}
\usepackage[width=122mm,left=12mm,paperwidth=146mm,height=193mm,top=12mm,paperheight=217mm]{geometry}
\usepackage{algorithm,algorithmic}
\usepackage{subfigure}
\newcommand{\etal}{\textit{et al}.}
\begin{document}

%\renewcommand\thelinenumber{\color[rgb]{0.2,0.5,0.8}\normalfont\sffamily\scriptsize\arabic{linenumber}\color[rgb]{0,0,0}}
%\renewcommand\makeLineNumber {\hss\thelinenumber\ \hspace{6mm} \rlap{\hskip\textwidth\ \hspace{6.5mm}\thelinenumber}}
%\linenumbers

\pagestyle{headings}
\mainmatter
\def\ECCV18SubNumber{****}  % Insert your submission number here

\title{An Aggressive Genetic Programming Approach \\ for Searching Neural Network Structure  \\ Under Computational Constraints} % Replace with your title

\titlerunning{Genetic Approach for Searching Neural Network Structures}

\authorrunning{Li, Xiong, Ren, Zhang, Wang, Yang}

%\author{Anonymous ECCV submission}
%\institute{Paper ID \ECCV18SubNumber}
\author{Zhe Li$^{1}$, Xuehan Xiong$^3$, Zhou Ren$^2$, Ning Zhang$^2$, \\ Xiaoyu Wang$^4$, Tianbao Yang$^1$\\
\{zhe-li-1,tianbao-yang\}@uiowa.edu, \\ \{zhou.ren, ning.zhang\}@snap.com, \\ \{xiong828, fanghuaxue\}@gmail.com}
\institute{$^1$The University of Iowa,$^2$Snap Research,$^3$Google Inc,$^4$Intellifusion}

\maketitle

\begin{abstract}
%Deep Convolutional Neural Networks (CNN) have achieved tremendous success for many computer vision tasks. However, a hand-crafted network structure tailored to one task may perform poorly on another task. Therefore,  it usually requires extensive amount of  human efforts to design an appropriate network structure for a certain task. 
Recently, there emerged revived interests of designing automatic programs (e.g., using genetic/evolutionary algorithms) to optimize the structure of Convolutional Neural Networks (CNNs)~\cite{lecun1998gradient} for a specific task. The challenge in designing such programs lies in how to balance between large search space of the network structures and high computational costs. Existing works either impose strong restrictions on the search space or use enormous computing resources. %In order to avoid high computational costs, many existing studies  usually impose strong restrictions on the search space (e.g., restricting the size of the network), which lead to unsatisfactory performance of the found networks. An exception to existing studies is a recent work~\cite{real2017large} that designed an automatic program using evolutionary techniques and running at unprecedented scales (e.g., hundreds  of GPUs, $\sim10^{20}$ floating-point operations). However, their approach is only affordable to giant industry companies with enormous computing resources. 
In this paper, we study how to design a genetic programming approach for optimizing the structure of a CNN for a given task under limited computational resources yet without imposing strong restrictions on the search space. To reduce the computational costs, we propose two general strategies that are observed to be helpful:  (i) aggressively selecting strongest individuals for survival and reproduction, and killing weaker individuals at a very early age; (ii)  increasing mutation frequency to encourage diversity and faster evolution.  The combined strategy with additional optimization techniques allows us to explore a large search space but with affordable computational costs. Our results on standard benchmark datasets (MNIST~\cite{lecun1998gradient}, SVHN~\cite{netzer2011reading}, CIFAR-10~\cite{krizhevsky2009learning}, CIFAR-100~\cite{krizhevsky2009learning}) are competitive to similar approaches with significantly reduced computational costs. %\ning{the last sentence is confusing, i thought you want to say even it is not better, we have lower computation cost than similar methods} %\textcolor{red}{***The abstract is too long. For CVPR paper, one half or two thirds of the column is the proper length.***}

\end{abstract}

\section{Introduction}
Although deep neural networks have achieved tremendous success in many domains (e.g., computer vision~\cite{Alexnet12,vggnet15,fastrcnn15}, speech recognition~\cite{hinton2012deep,dahl2012context}, natural language processing~\cite{dahl2012context,collobert2011natural}, games~\cite{silver2017mastering,silver2016mastering}), it still remains a great challenge to design the optimal  network structure for a certain task. Most existing works rely  on extensive human efforts on designing and experimenting with different structures.

\begin{figure}[t]
\centering
   \includegraphics[scale=0.45]{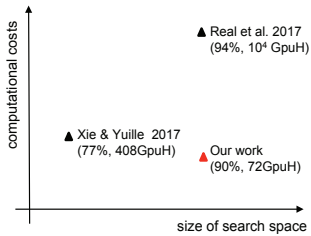}
\caption{A quick comparison with two recent similar works in terms of trade-off between the size of search space and computational costs. The first number in the parenthesis denotes the top 1 accuracy achieved on CIFAR-10 data and the second number denotes the number of GPU hours used. }
\label{fig:cs}
\end{figure}

Optimizing the network structures involves two fundamental issues:  how to define the search space of network structures; and how to design an efficient algorithm to search a good network structure in the search space. A great {challenge} in solving these issues lies {in} how to balance the {trade-off} between the size of search space and the computational {cost} of the {search} algorithm.  Earlier works based on \textit{neuro-evolution} for automatically discovering network structures usually impose strong restrictions on  the search space of the network structures due to limited computational power and scarcity of data~\cite{stanley2002evolving}.

%It's been witnessed that deep neural networks have already achieved tremendous success on a variety of computer vision tasks such as image classification \cite{Alexnet12,vggnet15,googlenet15,ResNet_cvpr16}, object detection \cite{RCNN14,sermanet2013overfeat,fastrcnn15,ren2015faster}, segmentation \cite{long2015fully,he2017mask}, video analysis \cite{xu2015discriminative,zha2015exploiting,gan2015devnet}, human pose estimation \cite{toshev2014deeppose}  among many others. The performance on these different tasks are dramatically boosted  by  sophisticated neural network structures such as AlexNet \cite{Alexnet12}, NIN (Network In Network) \cite{lin2013network}, VGG-Net \cite{vggnet15}, Inception Network \cite{googlenet15}, and ResNet \cite{ResNet_cvpr16}. However, designing these advanced neural network structure requires domain knowledge from experts and lots of effort.%-------------------------------------------------------------------------
%\\
Recently, there emerged revived interests in optimizing network structures (especially deep convolutional neural networks) using genetic/evolutionary algorithms~\cite{dufourq2017eden,xie2017genetic,real2017large}. However, the dilemma of computational costs and search space {trade-off} has pushed these works into two ends. At one end, one has to restrict the search space by imposing strong constraints on the network structures. For example, in \cite{xie2017genetic} a network is composed of a fixed number of stages and each stage is composed of a fixed number of nodes representing convolutional operations. In~\cite{dufourq2017eden}, {Dufourq and Bassett} restricted mutation operations to adding, deleting and replacing a randomly selected layer in a network with a predetermined maximum number of layers (e.g., $7$ is used in their experiments). As a result, their evolved networks share a single path structure in contrast to a multiple-path structures as in residual networks~\cite{ResNet_cvpr16} and Inception~\cite{googlenet15}. On the other end, {Real~\etal}~\cite{real2017large} investigated large-scale evolution of networks operating at unprecedented scale by using a large amount of computing resources (e.g., spending over 10 days on  250 GPUs), which verifies neuro-evolution can achieve competitive performance as  hand-crafted models built on many years of human experience. However, such an brute-force approach is not affordable for general users who have limited computational resources. 

In this paper, we focus on optimizing deep CNN {structures} for image classification due to the availability of existing results for comparison and its popularity in computer vision.  Similar to~\cite{real2017large}, we also study evolution-based algorithms, which search CNNs in a large search space that is defined by a set of mutations. Nevertheless, the difference from previous works~\cite{dufourq2017eden,xie2017genetic,real2017large} is that  our main focus is to tackle  an important and challenging question for optimizing neural network structures, i.e.,  {\bf how to maximize the exploration in the search space under limited computational resources~\footnote{computational resources include not only  the hardware but also the computing time.}}. Instead of imposing strong restrictions on the search space, we propose new effective strategies to reduce the computational costs. We use an aggressive method to select strong individuals for survival and reproduction. In particular, among a set of individuals (i.e., population)  only  a small number of fittest individuals that are sufficiently different from each other are selected for producing the next generation. This strategy avoids wasting time on training weaker individuals that may eventually be eliminated in a later stage. However, a potential issue caused by this strategy is that the diversity of population decreases, which is very important for genetic programming. To remedy this issue, we propose to (i) increase the number of possible mutations; (ii) make clones of the selected fittest individuals to undergo different mutations.  Additional  techniques  are also investigated to speed up the search process and to shorten the training time of each individual during evolution. Before ending this section, we give a quick comparison of our work and two recent works highlighting the trade-off between the size of search space and computational costs in Figure~\ref{fig:cs}.

The main contribution of this paper can be summarized as following: 1) Through extensive experiments, we show that we can automatically (without any human effort at all, for example, tuning, modification, adding other layers) design network structures to achieve competitive performance, which can be conducted by general users with limited computing resources; 2) From empirical experiments, we found some interesting insights in designing neural networks, for example, skip layers was found in early stage of searching process later on was replaced by other layers; 3) We propose a simple yet efficient selection strategy, which performed better compared to other strategies, has its interest and can easily be adopted by practitioners.

\section{Related Work}
%In this section, we review some related works on automatically designing deep neural network structures.

There exist abundant studies on  using genetic or evolutionary algorithms for discovering neural networks structures before the re-emergence of deep learning~\cite{miller1989designing,stanley2002evolving,gruau1993genetic} in 2012. These algorithms are also known as \textit{neuro-evolution}. Most of these works  are restricted to feedforward neural networks of a few layers. However, many techniques in these earlier works are also useful for optimizing large convolutional neural networks.  In designing a neuro-evolution algorithm, several fundamental questions need to be answered, including (i) how to encode an individual; (ii) what are the allowed mutations; (iii) how to select individuals for reproduction; (iv) what is the fitness function. Existing works may differ from each other on how to address these questions, which are also related to a fundamental issue in neuro-evolution (and also in other  meta-heuristic optimization algorithms): how to balance between the size of search space and the computational costs. In the following discussion, we will highlight how to address these fundamental questions and how to  balance the trade-off. 

%\subsection{Designing deep neural networks} 
%The widely used neural network structures such as AlexNet \cite{Alexnet12}, NIN (Network In Network) \cite{lin2013network}, VGG-Net \cite{vggnet15}, Inception Network \cite{googlenet15}, and ResNet \cite{ResNet_cvpr16} are handcrafted or modified from others, even though automatically designing deep neural network structures could date back to work \cite{schaffer1992combinations}, in which they tried to aumatically find structure and weight of model.

%\subsection{Genetic algorithm}
%Genetic algorithms are the type of optimization and search methods, which have been successfully used a varities of areas such as music generation and so on. An entire process of applying genetic algorithm to specific domain includes: 1) encoding solution of interested problem to some format of chromosome; 2) evulating fitness score of individual solution; 3) selecting survived solutions; 4) defining mutation and crossover operations. 

%\subsection{Designing deep neural networks with genetic algorithms}
In~\cite{xie2017genetic}, the authors developed a genetic algorithm using a fixed-length binary
string to encode the network structure. The search space consists of all networks with a fixed number of stages, where each stage is composed of a fixed number of nodes representing convolutional operations. Mutations are easily operated by randomly flipping each bit in the string representation, which correspond to adding, deleting and changing connections of nodes within each stage.  By restricting the number of stages (e.g., 3) and the number of nodes in each stage, their computational cost is controlled under a manageable level. The selection of individuals for reproduction is done by a Russian roulette process, which selects individuals based on a non-uniform distribution whose probabilities are proportional to the fitness of the individuals, i.e., individuals with higher fitness score will be selected with higher probability. 

The focus of~\cite{real2017large} is to scale up neuro-evolution to take advantage of the tremendous computational resources Google LLC has. A total of 250 GPUs are used in their experiments. They used a graph to encode an individual, and defined seven mutations to change the structure of networks~\footnote{they also used several other mutations that do not affect the structure of the network.}, and used a standard binary tournament selection method~\cite{goldberg1991comparative} for selecting individuals for reproduction. 

The fitness function of the above works purely depends on the performance of individual models on a validation data, which are trained by back-propagation. The evolutionary techniques used in~\cite{dufourq2017eden}
is similar to that in~\cite{real2017large} except that their mutations are restricted to adding, deleting and replacing one of six predefined layers, which include two-dimension convolution, one-dimension convolution, fully connected, dropout, one-, and two-dimension max pooling. As a result, their algorithm cannot discover multiple-path networks, which are prevalent in modern deep learning community. A common feature of these works is that they use a combined strategy that lets the structure evolve but optimizes the weights of each individual by back-propagation, which is also adopted in the present work.

Another fundamental issue in genetic/evolutionary  algorithm is the diversity of the population. A traditional approach for encouraging diversity of population is fitness sharing, where the fitness of each individual is scaled based on its proximity to others. It means that originally good solutions in densely populated regions will be given a lower fitness value than comparably good solutions in sparsely populated regions. All the three  recent works~\cite{xie2017genetic,real2017large,dufourq2017eden}  did not use any type of fitness sharing to encourage diversity. In~\cite{real2017large}, the authors simply use a very large population size (i.e., 1000) to increase the diversity. A key difference between our work and these previous work lies in the selection process and the number of mutation operations. The proposed solution takes both the limitations of computational resources and diversity of the population into account. As a result, even though we do not impose any strong restriction on  the search space, we can use less computational costs to achieve competitive if not better prediction performance than~\cite{xie2017genetic,dufourq2017eden,real2017large}.

Other related works on automatically discovering network structures include reinforcement learning~\cite{baker2016designing,zoph2016neural} based approaches and Bayesian optimization~\cite{snoek2012practical,bergstra2013making,mendoza2016towards} based approaches. We refer the readers to~\cite{real2017large} for more discussion and references.  
\section{Our Approach}
The proposed algorithm follows the standard flow of neuro-evolution, i.e., population initialization, individual selection, reproduction, mutation/cross-over, and fitness score evaluation. The individuals in the initial population are simple neural network structures with only one global pooling layer or one fully connected layer.  We use an acyclic  graph to encode an individual with each node in the graph representing a basic operation or connection including \textit{convolution, pooling, fully connected, concatenation and skip}.  Those operations are standard in the neural network literature. Please refer to Figure~\ref{fig:examples_mutation_operations} for examples of individuals represented by a graph. We also use the prediction performance on  a validation data as fitness score.  %\ning{should we mention the size/flops constraint here?}\textcolor{red}{I don't think it is necessary, even though it is implemented in library but it seems not related}

However, different from~\cite{xie2017genetic,real2017large,dufourq2017eden}, we are not only exploring how to implement a genetic/evolutionary algorithm for optimizing deep convolutional neural networks, but also exploring how to  reduce the computational costs under the framework of neuro-evolution without imposing strong restriction on the search space. Next, we will  present our strategies for reducing the computational costs.

\begin{figure*}[t]
\begin{center}
      \includegraphics[scale=0.34]{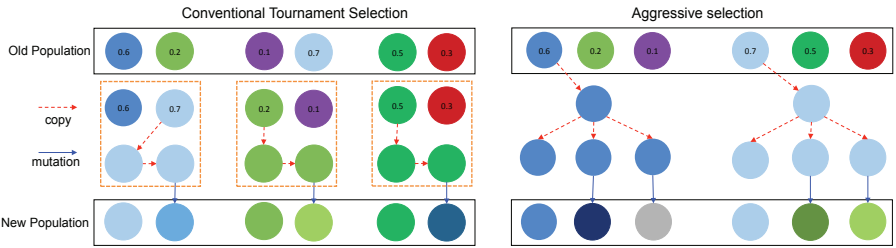}
   
\end{center}
\caption{Comparison between the proposed aggressive selection and mutation strategy (right) vs conventional tournament selection and mutation strategy (left). Each colored ball denotes an individual, the number within each ball denotes its fitness score, red dashed arrows denote a copy and green solid arrows denote mutations.}
\label{fig:top2_selection}
\end{figure*}

\begin{algorithm}[t]
\caption{Aggressive Selection of top-$k$ individuals}\label{alg:agg}
\begin{algorithmic}[1]
\STATE \textbf{Input}:  a population of individuals ranked according to their fitness score from large to small, $\mathcal P_{t-1}=\{i_1, \ldots, i_N\}$. A target number of individuals $k$ and a distance threshold $d$.
\STATE Initialize an empty set $\mathcal P_t$
\FOR{$j=1,2,\ldots,N$}
\STATE choose the next individual $i_j$ in $\mathcal P_{t-1}$
\IF{the distance between $i_j$ and individuals in $\mathcal P_t$ exceeds a certain threshold $d$ }
\STATE  add $i_j$ into $\mathcal P_t$
\ENDIF
\IF{the size of $\mathcal P_t$ is equal to $k$}
\RETURN $\mathcal P_t$
\ENDIF
%\ELSE

%Find a vector $\v_j$ such that $\|\v_j\|=1$ and, with probability at least $1-\delta'$,
%\[
%\lambda_{\min}(\nabla^2 f(\x_j))\geq \v_j^{\top}\nabla^2 f(\x_j)\v_j - \max(\epsilon_2, \|\nabla f(\x_j)\|)/2
%\]
%using a leading eigen-vector computation. 
%\STATE 
\ENDFOR
\end{algorithmic}
\end{algorithm}

\subsection{Aggressive selection and mutation}
A potential issue in traditional selection strategies (e.g., tournament selection or sampling-based selection) is that weak individuals might survive for a long period. While this feature is helpful to increase the diversity of the population, however it may waste a lot of time to train these weak individuals that  will eventually be eliminated. We propose to eliminate these weak individuals at their very early age, and use other approaches to increase the diversity of the population. The algorithmic description of the proposed aggressive selection is presented in Algorithm~\ref{alg:agg}, and an illustration of the proposed selection process is presented in Figure~\ref{fig:top2_selection}.  In particular, we greedily select the top $k$ individuals from a population of individuals $\mathcal P_{t-1}$ based on their fitness scores. To encourage the diversity, we also make sure the distance between the selected top individuals exceeds a certain threshold. The distance between two individuals is computed by comparing the nodes of two graph structures from input layer to output layer. If the nodes are represented by an alphabet denoting their operation or connection types, the distance is simply the hamming distance. It is notable that the distance between individuals is also considered by fitness sharing in previous studies~\cite{goldberg1991comparative} to encourage diversity. 

\paragraph{Multiple Cloning.} With the aggressive selection strategy described above, we can eliminate many weak individuals at their early ages. However, the small number of retained individuals will reduce the size of the next population and thus restrict diversity. To address this issue, we will resort to cloning, i.e., making multiple copies of the selected individuals to undergo different mutations for generating the next population. For comparison, in traditional tournament selection as illustrated in Figure~\ref{fig:top2_selection}, a weak individual might be selected and each survived individual only undergoes one mutation, which is the strategy adopted in~\cite{real2017large,dufourq2017eden}. In conventional sampling-based selection and mutation, each individual has a certain probability to be retained and the retained individual has a certain probability to be mutated. This is the strategy adopted in~\cite{xie2017genetic}, where the mutation probability is set to a small value (e.g. 0.05). 
We can see that the proposed selection and mutation strategy is more aggressive than the existing works in that only a small number of strong individuals are retained and each survived individual reproduces themselves to undergo more mutations for potential growth in the fitness. 

%\subsection{Crossover/mutation}
\subsection{Mutation operations}
\label{sb:mutation-operation}
\begin{figure}[t]
\begin{center}
   \includegraphics[scale=0.40]{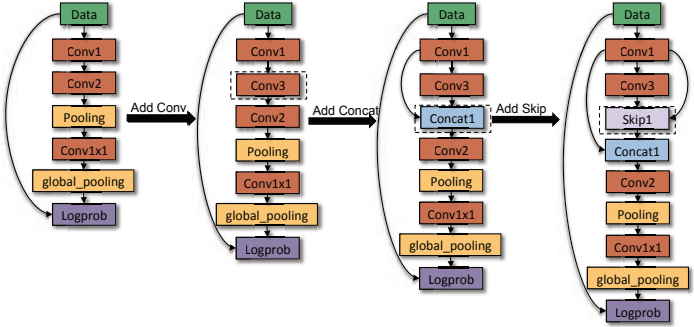}
\end{center}
\caption{Example of how a neural network structure  mutates to new ones after undergoing different mutation operations: \texttt{add\_convolution}, \texttt{add\_concatenate} and \texttt{add\_skip}. The dotted squares mark added convolution layer ``Conv3'', concatenate layer ``Concat1'' and skip layer ``Skip1''.}
\label{fig:examples_mutation_operations}
\end{figure}

To complement our proposed aggressive selection and mutation, we increase the number of possible mutation operations compared to the existing works ~\cite{real2017large,dufourq2017eden,xie2017genetic}. We define 15 different types of mutation operations as shown in Table \ref{tb:implemented_mutation_operation_list}, which almost doubles the amount considered in~\cite{real2017large}. Note that three mutation operations (\texttt{reset\_weight}, \texttt{continue\_training}, \texttt{alter\_learning\_rate}) appeared in \cite{real2017large} are not included in Table \ref{tb:implemented_mutation_operation_list} since those operations do not change the structure of a neural network. Next, we provide more details regarding each mutation operation. 
% Clearly, we define more mutation operations, which helps enlarge neural network structure search space compared to that in previous work \cite{real2017large}. This enlarged set of mutation operations on other side allows us adopt aggressively selection strategy (choosing top 1 or 2 strongest neural network structure to survive and reproduce) without sacrificing diversity of population for exploration. 
In the following, we discuss some implementation details of each mutation operation. We are going to focus on the \texttt{add} operations. For all \texttt{removal} operations, we randomly select and remove one of existing layers of the chosen type. If no such layer exists, no operation will be applied.
%\ning{should we change table 1 font too?}
\begin{itemize}
\item \texttt{add\_convolution}: Firstly, we randomly select the position to add a convolution layer. Then we insert a convolution layer with channel number $32$, stride $1$, filter size $3 \times 3$, and number of padding pixel $1$. For simplicity, those values are chosen to ensure the input and output of the feature map dimensions do not change after the convolution. Note that even though we use a predefined set of channel number, stride and filter size, those values could be altered later through \texttt{alter\_channel\_number}, \texttt{alter\_stride}, \texttt{alter\_filter\_size} mutation operations, which we will discuss shortly. In this work, a convolutional layer is by default followed by batch normalization~\cite{ioffe2015batch} with Relu~\cite{Alexnet12} activation unit. 

\item \texttt{alter\_channel\_number, alter\_stride, alter\_filter\_size}: These three types of mutation operations are to reset the hyper-parameters in a convolution layer. We randomly choose a new value for the corresponding hyper-parameter from a predefined list, i.e., $\{8,16,32,48,64,96,128\}$ for  channel numbers, $\{1\times 1, 3 \times 3, 5 \times 5\}$ for filter sizes, and $\{1,2\}$ for strides. 

\item \texttt{add\_skip}: A skip layer, illustrated in Figure~\ref{fig:examples_mutation_operations}, is to implement the skip connection introduced in residual networks~\cite{ResNet_cvpr16}. Since a skip layer requires its two bottom layers to share the same feature map dimension and channel number,  we first find out all pairs of layers which could potentially be bottom layers of the skip layer. Then, a skip connection is added on top of a randomly selected pair from all possible pairs.

\item \texttt{add\_concatenate}: Similar to skip layer, a concatenate layer requires two bottom layers to share the same feature map dimension, but they could have different channel numbers. Thus, the \texttt{add\_concatenate} mutation  follows a similar procedure as \texttt{add\_skip} mutation.

\item \texttt{add\_pooling:} Here, we restrict the insertion of a pooling layer such that it can only take place right after a convolution layer. For simplicity, we limit the pooling strategy to be max pooling and kernel size to $2\times2$ with stride $2$. This predefined pooling configuration could be relaxed in future work.

\item \texttt{add\_fully\_connected}:
For this operation, we limit its position to be the last layer or immediately following another fully connected layer. The output dimension of this inserted layer is uniformly chosen from the following set $\{50,100,150,200\}$.

\item \texttt{add\_dropout}: For this operation, we limit its position to be immediately after a fully connected layer. For simplicity, we set dropout ratio to $0.5$. %For removing mutation, we also uniformly decide which dropout layer should be removed. 
\end{itemize}

Note that applying some mutation operations (e.g., \texttt{alter\_stride}, \texttt{add\_pooling}) may result in inconsistency of feature map dimensions. In this situation, we will adopt the following strategies to address the above issue: i) adding additional padding pixel; ii) adding a $1 \times 1$ convolution layer to adjust channel numbers. If it still results in an invalid network structure, we simply apply a mutation again. 

In Figure~\ref{fig:examples_mutation_operations}, we show an example of how a neural network structure evolves to new ones after undergoing some mutation operations: \texttt{add\_convolution, add\_concatenate, add\_skip}. We use the dotted square to mark the new layers. Clearly, we could see that as network evolves, we can explore the diverse neural network structures and obtain neural network structure with potential better performance. 

\begin{table}[]
\begin{center}
\begin{tabular}{l|l|l}
\hline
Mutations           & \cite{real2017large} & Ours \\ \hline \hline
\texttt{add\_convolution}    & \checkmark  & \checkmark     \\ \hline
\texttt{remove\_convolution}  & \checkmark & \checkmark     \\ \hline
\texttt{alter\_channel\_number}     &\checkmark          &\checkmark \\ \hline
\texttt{alter\_filter\_size}      &\checkmark      &\checkmark      \\ \hline
\texttt{alter\_stride}        &\checkmark          &\checkmark      \\ \hline
\texttt{add\_dropout}        & -        &\checkmark    \\ \hline
\texttt{remove\_dropout}     & -        &\checkmark    \\ \hline
\texttt{add\_pooling}                  & -        &\checkmark     \\ \hline
\texttt{remove\_pooling}               & -        &\checkmark      \\ \hline
\texttt{add\_skip}                     &\checkmark &\checkmark      \\ \hline
\texttt{remove\_skip}                  &\checkmark &\checkmark      \\ \hline
\texttt{add\_concatenate}        &  -       & \checkmark    \\ \hline
\texttt{remove\_concatenate}     &  -       & \checkmark    \\ \hline
\texttt{add\_fully\_connected}    &  -        &\checkmark      \\ \hline
\texttt{remove\_fully\_connected} &  -        &\checkmark      \\ \hline
\end{tabular}
\end{center}
\caption{The allowed mutation operations in our work and in~\cite{real2017large};~\checkmark represents that mutation operation is defined while - represents not available}
\label{tb:implemented_mutation_operation_list}
\vspace{-0.25in}
\end{table}

%\begin{table}[h]
%\begin{center}
%\begin{tabular}{|l|l|l|}
%\hline
%add convlution       & remove convolution   & add pooling        \\ \hline
%remove pooling       & add dropout          & remove dropout     \\ \hline
%add fully connect    & remove fully connect & add skip           \\ \hline
%remove skip          & add concat      & remove concat \\ \hline
%alter channel num & alter filter size    & alter stride       \\ \hline
%\end{tabular}
%\end{center}
%\caption{List of Mutation Operations implemented}
%\label{tb:list-mutation-operations}
%\end{table}

%Crossover is operation that is applied to two neural network structures(father and mother) to get their two children. Specifically, given father and mother neural network structures, genetic algorithm first find out all pair of proper crossover points between father and mother neural network structures. The proper crossover points refer to layers that dimension are matchable in father and mother neural network structure. Then aglorithm uniformly chooses one pair crossover points to swap part neural network structure after crossover points between father and mother neural network structures. 
\subsection{Training strategy}
\label{sb:aggressive-training}
To evaluate the fitness score of each individual, a standard approach is to use existing optimization algorithms off-the-shelf to learn the weight parameters. Training a CNN may take tens of thousands gradient descent iterations to achieve a good local minimum. However, we observe that ``deep'' training (i.e., setting a very stringent condition for stopping the training process) is not necessary during the evolution since our goal is to have a ranked list of individuals for selection. Therefore, a rough estimate of the prediction performance for each individual is sufficient for driving the evolution. %\ning{i feel this therefore claim is not valid}

To reduce the training time of each individual during the evolution, we explore a different learning rate decay strategy to train a deep neural network.  There are two popular strategies for decaying the learning rate. One method is an inverse learning rate decay strategy~\cite{jia2014caffe}, where the learning rate $\eta_t$ at the $t$-th iteration is set to 
%It is notorious that utilizing genetic algorithm for searching optimal neural network structures requires unprecedented scale of computational resource. Exploring strategies to reduce computational costs at same time without sacrificing performance becomes more and more crucial. In this subsection, we present several general strategies for reducing computational costs. 
%\begin{itemize}
%\item 
%To what extend to train each neural network structure: Hand training one neural network usually requires large number of iterations to update model to achieve better performance. However, it is not computationally affordable in genetic programming due to massive  population of individual neural network structure along evolution process. Thus, it is demanded to aggressively explore potential performance of individual neural network in small number of iterations. For this purpose, we only train each neural network with maximum $20$k iterations with proposed learning strategy called mutli-inverse learning rate. We divide $20$k iteration in three stage, first $10$k iterations, $10$k to $15$k iterations and finally $5$k iterations. In each stage we adopt inverse learning rate strategy as 
\begin{equation}\label{eqn:lr}
\eta_t = \eta_0*(1 + \gamma*t)^{-\alpha}
\end{equation}
where $\eta_0$ is the initial step size and $\gamma, \alpha$ are the hyper-parameters. Another popular method is a multi-stage strategy~\cite{Alexnet12}, where the learning rate is reduced by a fixed factor (e.g., 10) after a large number of iterations. We use a mixture of both strategies. We divide our training process into three stages with a maximum number of $20000$ iterations: with the first stage being the first $10000$ iterations, from $10000$ to $15000$ iterations as the second stage, and the last $5000$ iterations as the final stage. Within each stage, we use the inverse learning rate strategy. The learning rate is reduced by a fixed factor after each stage. This strategy avoids running a large number of iterations with the same step size without improving the prediction performance much at each stage,  and also quickly gives a rough estimate of the prediction performance without spending long time at the tail of the learning curve that has little improvement on the prediction performance. Finally, after the neuro-evolution process terminates with a good structure, we switch to existing optimization algorithms for deep training.%\ning{does it mean we train longer for the selected structure? isn't that clear}
\vspace{-0.12in}
\subsection{Mutation operation sampling} 
To speed up the evolution process, we also use non-uniform sampling probabilities for choosing a mutation operation. Using uniform probabilities to choose a mutation operation will waste lots of time training weak individuals that are mutated by removing convolution, skip, concatenation from their parents in the early stage of evolution process. 
%We explore which mutation operations should be applied more than others to achieve better performance in order to reduce computation cost. 
To avoid this issue,  we explicitly set the sampling probabilities of \texttt{add\_convolution}, \texttt{add\_skip}, \texttt{add\_concatenate}, \texttt{alter\_ stride} \texttt{alter\_filter\_size}, and \texttt{alter\_channel\_number} two times larger than that of other mutation operations at the earlier stage of the evolution process.  %\ning{change font too also in figure 3 caption}\textcolor{red}{done}

\section{Experiments}
In this section, we report some experimental results of the proposed aggressive genetic programming approach for optimizing convolutional neural network structures. We emphasize that we are not aiming to achieve better performance than~\cite{real2017large} due to limited computing resources. Instead, we focus on showing that the proposed strategies can reduce the computational time and also  achieve competitive and even better performance than similar works using significantly less computational power. 

In subsection~\ref{sec::exp::config}, we describe the experiment setup including datasets and data preprocessing. In subsection~\ref{sec::exp::selection}, we show the performance of the aggressive selection strategy under different values of $k$ on CIFAR-10~\cite{krizhevsky2009learning} dataset to justify the proposed aggressive selection. In subsection~\ref{sec::exp::comparison}, we compare the performance of the proposed aggressive evolution with  other genetic approaches as well as previous works that achieved the-state-of-art results by hand-crafted networks on four standard benchmark datasets. In subsection~\ref{sec::exp::evolve}, we present the discovered neural network structures for CIFAR-10 and CIFAR-100.% that are different from human-designed ones.   
%Last, we present the results on the learned neural network structures if introducing number of parameters and float point operations on CIFAR10 dataset. 
%Before we present those three part experimental results, we give brief explanation of dataset, experiment configuration. %, and hype-parameters of training each individual neural network structures.
\subsection{Experimental setup}
\label{sec::exp::config}
\noindent{\textbf{Datasets and Preprocessing}}: We conduct experiments on four benchmark datasets: MNIST~\cite{lecun1998gradient}, SVHN~\cite{netzer2011reading}, CIFAR-10~\cite{krizhevsky2009learning} and CIFAR-100~\cite{krizhevsky2009learning}. MNIST dataset contains $60,000$ training images and $10,000$ test image where each gray-scale image contains one of the $10$ digits, $0$ to $9$. CIFAR-10 dataset \cite{krizhevsky2009learning} has 50,000 training images and 10,000 test images. It contains 10 classes and each RGB image has a size of $32 \times 32$. The data is preprocessed by applying a Global Contrast Normalization (GCN) and ZCA whitening \cite{goodfellow2013maxout} and each side is padded with four pixels. In the training phase, a $32\times 32$ patch is randomly cropped from the padded image while in the test phase the original images are used. CIFAR-100 dataset is similar to CIFAR-10 dataset but has 100 classes in total. SVHN is a street view house number dataset which contains about $73,257$ training images and $26,032$ test images.  
\\

\noindent{\textbf{Experiment configuration:}} In our experiment, the population size is set to be 10. Given a population of $10$ individuals (which are clones of top $k$ individuals in intermediate generations), we let each individual undergo a mutation, and then select top $k$ individuals from the 10 mutated individuals and the original $10$ individuals. A new population will be created by making equal number of clones of the selected individuals to reach the population size $10$. 
It is worth mentioning that even though we use such a small population size, our performance is competitive and even better than~\cite{dufourq2017eden,xie2017genetic}, in which the population size is set to  100 and 20, respectively. It is expected that using a larger population size will further increase our performance according to~\cite{real2017large}. We use mini-batch Stochastic Gradient Descent (SGD) to train each individual neural network for a maximum of 20,000 iterations with a momentum 0.9. The mini-batch size is fixed to be 128. The weight decay is set to be 0.0005. The learning rate strategy is described in subsection~\ref{sb:aggressive-training}. The initial learning rates for the three stage are set to be $10^{-1}, 10^{-3}, 10^{-5}$, respectively. The parameters in~(\ref{eqn:lr}) are set to $\gamma = 0.001$ and $\alpha = 0.75$. The distance threshold in aggressive selection is set to be $1$. In our experiments, one evolution process is always run on one GPU. 
%\vspace{-0.3in}
\subsection{The effect of aggressive selection}
\label{sec::exp::selection}
\begin{figure}[t]
 \centering
   %\subfigure{\includegraphics[scale=0.34]{figures/cifar10_acc_best_worst_top1_top2_top5_top10}}%\hfill
    \subfigure{\includegraphics[scale=0.15]{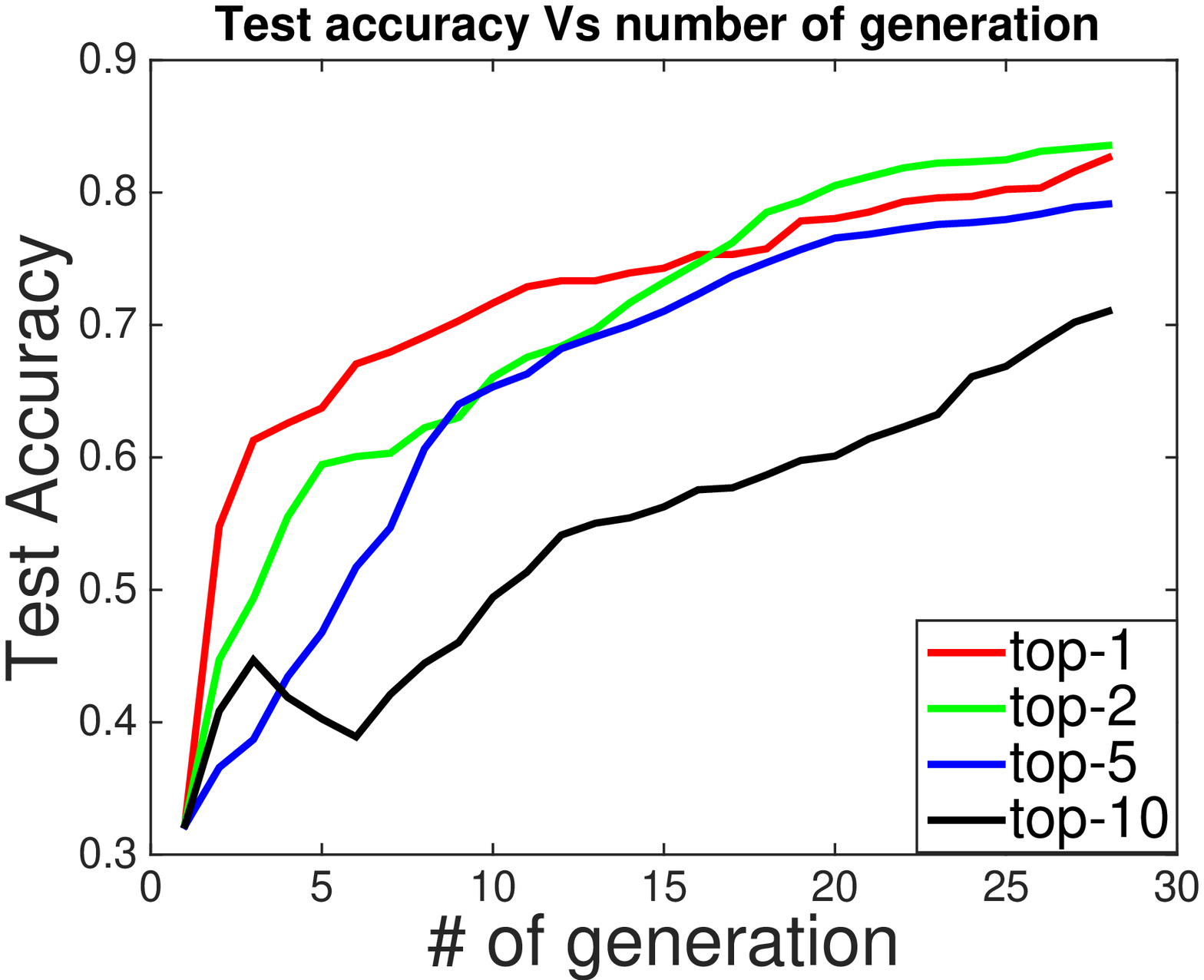}}
    \subfigure{\includegraphics[scale=0.11]{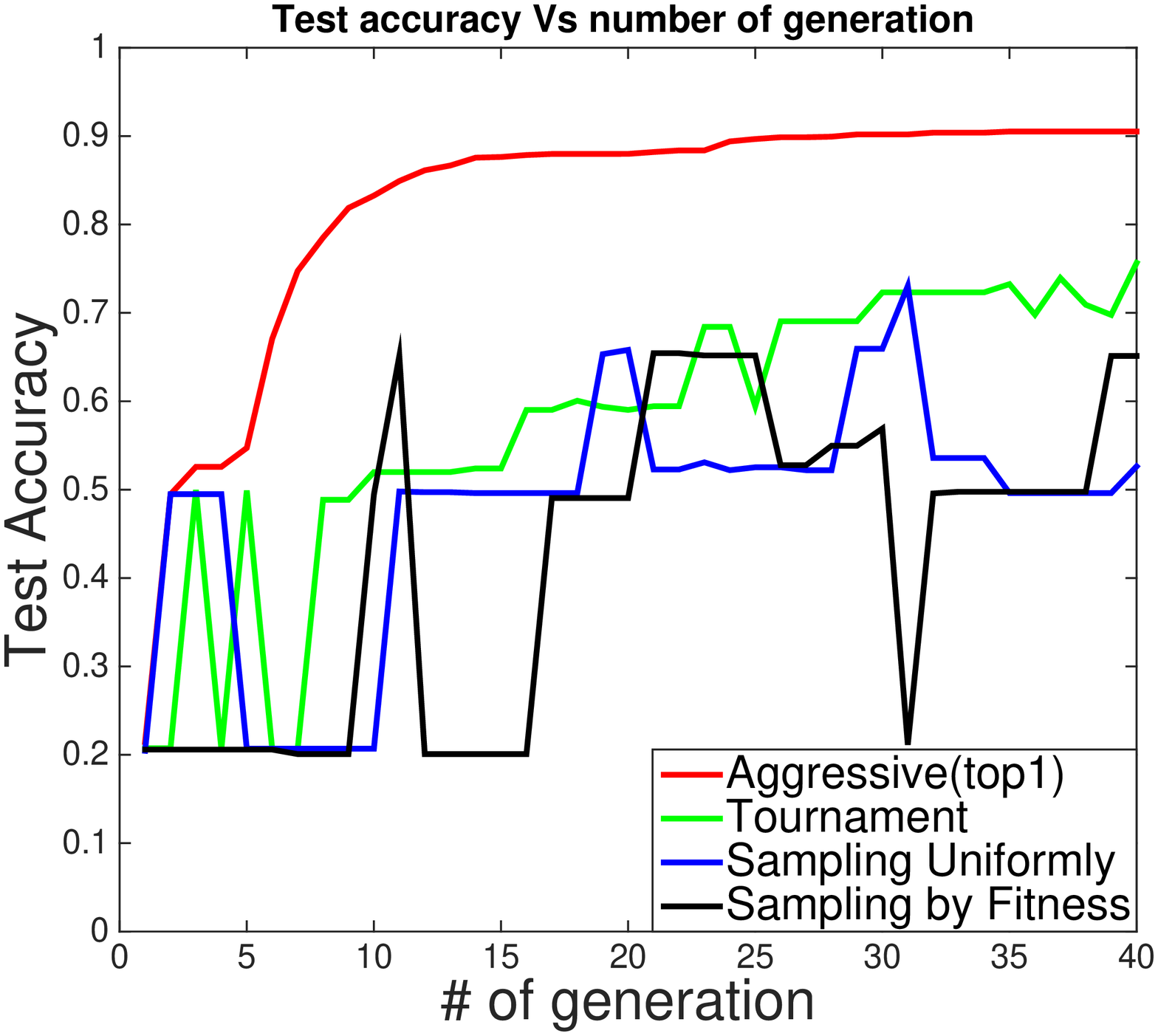}}
     \subfigure{ \includegraphics[scale=0.115]{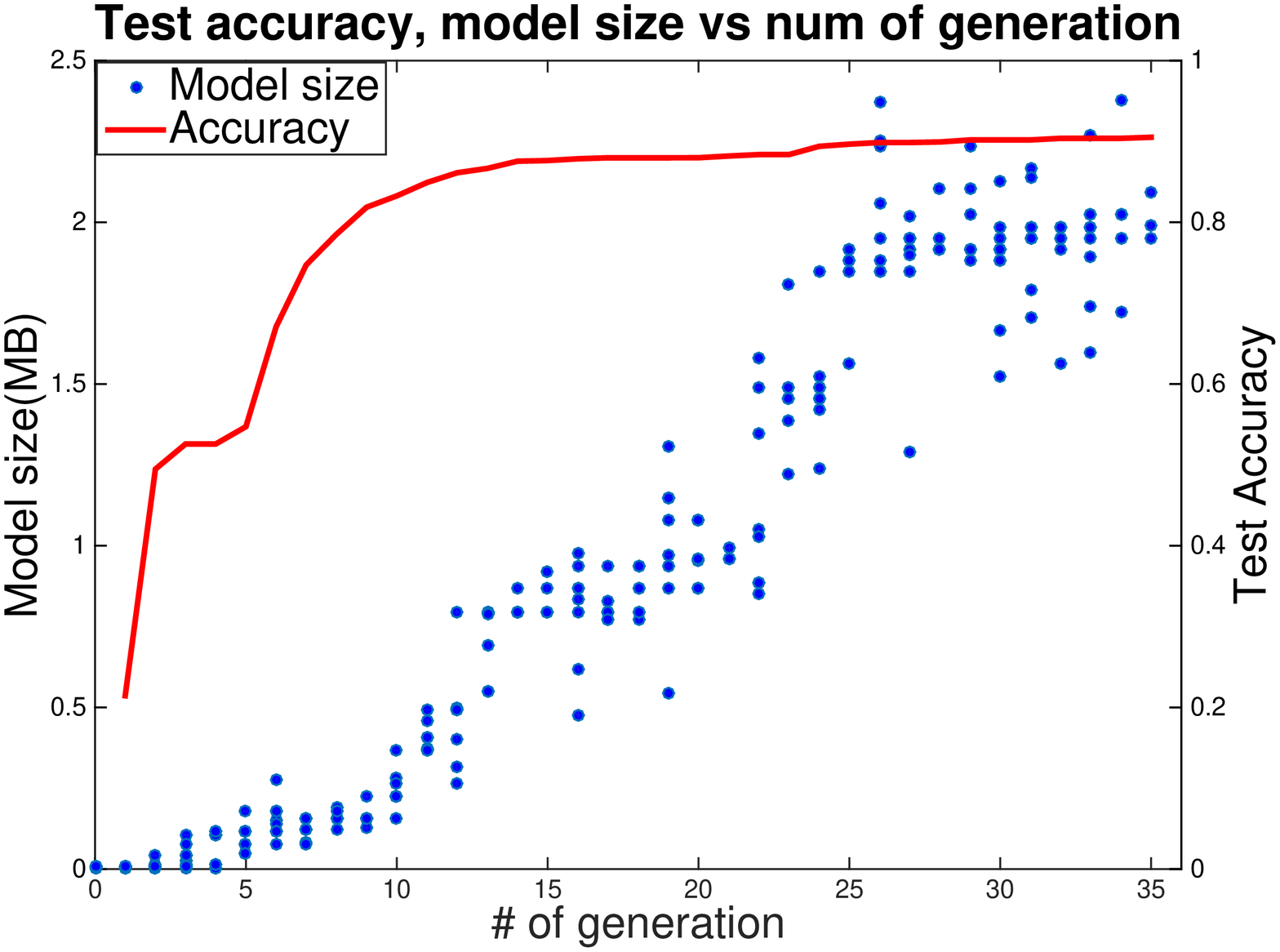}}
    %\subfigure{ \includegraphics[scale=0.16]{figures/cifar10_model_size_acc_v0}}
    %\subfigure{   \includegraphics[scale=0.20]{figures/top1_best_worst_bar.eps}}
    %\hfill
\caption{\textbf{Left}: The test accuracy of the best individual among the selected  top $k$ individuals vs the number of generations  on CIFAR-10 dataset, where different curves correspond to  aggressive selection with different values of $k$. \textbf{Middle}: The test accuracy of the best individual among the selected individuals by using aggressive, tournament, sampling uniformly and sampling by fitness selection strategies vs the number of generations  on CIFAR-10 dataset. \textbf{Right}: The evolution of model size and test accuracy of the best individual in \textbf{AG-Evolution} algorithm on CIFAR-10.
}
\label{fig:cifar10-acc-topk}
\end{figure}

Here, we present the evidence that the proposed aggressive selection strategy can dramatically speed up the evolution process. The following experiments are conducted in the CIFAR-10 dataset. In the left of Figure~\ref{fig:cifar10-acc-topk}, we plot the evolved network performance under four different values of $k=1, 2, 5, 10$ used in our aggressive selection strategy. The smaller the $k$ value, the more aggressive the selection strategy. For each experiment setting, we plot the test accuracy of the best individual among the selected top $k$ individuals from each generation.  We observe that aggressive selection with smaller values of $k$ (e.g.,  $1$ and $2$) evolves  faster than non-aggressive selection using larger values of $k$ (e.g., $5, 10$). %For the most aggressive setting ($k=1$), we also plot the performance of  the best individual and worst individual from the population before selection  at 5 generations in Figure~\ref{fig:cifar10-acc-topk}. We can see that even we aggressively select the top $1$ individual, after mutations we still have diverse individuals. 
We further compare the proposed aggressive selection strategy with other existing selection strategies such as \textbf{Tournmanet}, \textbf{Sampling Uniformly} and \textbf{Sampling by Fitness}. In the middle of Figure ~\ref{fig:cifar10-acc-topk}, we plot the test performance of the best individual in one generation by using those different selection strategies, from which we can observe that aggressive selection evolves faster than other strategies dramatically. For researchers and practitioner in genetic programming, this proposed competition strategy is of interest and can be easily adopted.

%-------------------------------------------------------------------------
\subsection{Comparison with existing methods}
\label{sec::exp::comparison}
In this section, we compare the performance of the proposed aggressive genetic programming approach with existing genetic approaches on benchmark datasets MNIST, SVHN, CIFAR-10, CIFAR-100.  We refer to the genetic approaches presented in~\cite{dufourq2017eden,xie2017genetic,real2017large} as \textbf{EDEN}, \textbf{Genetic-CNN} and \textbf{LS-Evolution}, respectively. For our method, we report the result using aggressive selection with $k=1$, referred to \textbf{AG-Evolution}. For reference and comparison, we also include  state-of-the-art results based on hand-crafted neural network structures (referred them as \textbf{SOTA}) as well as the results using non-aggressive selection (i.e., by setting $k=10$ in our framework), which is referred to as \textbf{NA-Evolution}. 

For \textbf{NA-Evolution} and \textbf{AG-Evolution}, we terminate the evolution process when the performance on the validation data saturates on all datasets except on the CIFAR-100 dataset, in which we terminate the process earlier. It is possible that by continuing the evolution process, the performance might be further improved.   We would also like to emphasize that state-of-the-art results could be attributed to not only a good network structure but also some other factors (e.g., using a good pooling function~\cite{lee2016generalizing}), which are not considered in current genetic approaches. For \textbf{Genetic-CNN}, we do not compare with the results in their Table 3 for SVHN, CIFAR-10 and CIFAR-100. The reason is that their results in Table 3 are not directly achieved by the evolution process. They re-trained the networks by using large number of filters, which are not exactly the networks found by their genetic algorithm. 

%, we compare test accuracy of  the final best neural network structure generated by our genetic approach, the hand-crafted state-of-the-art neural network structures (refereed to SOTA), and neural network structures found by other existing genetic approaches on benchmark datasets MNIST, SVHN, CIFAR-10, CIFAR-100. 
Table~\ref{mnist-quantitive-comp},~\ref{svhn-quantitive-comp},~\ref{cifar10-quantitive-comp},~\ref{cifar100-quantitive-comp} show our results on different datasets. For each method, we report both the test accuracy and the computational cost measured by the total number of used GPU hours (GPUH), which is used in \cite{xie2017genetic,dufourq2017eden,real2017large}. The GPUH numbers for other methods are directly from the original papers. It is notable that on some datasets the results of  \textbf{EDEN}, \textbf{Genetic-CNN} and \textbf{LS-Evolution} are missing, which is because they are not reported in the original papers. From the results, we have the following observations. \begin{itemize}
\item First, the performance of the discovered neural network structures by our genetic approach \textbf{AG-Evolution} on MNIST and SVHN datasets is very close to  the state-of-the-art results. 
\item Second, compared with \textbf{EDEN}~\cite{dufourq2017eden}, \textbf{AG-Evolution} achieves  better performance on MNIST and CIFAR-10, and compared with \textbf{Genetic-CNN}~\cite{xie2017genetic}, \textbf{AG-Evolution} can find a much better neural network on CIFAR-10 ($0.9052$ vs $0.7706$ for test accuracy) with much less time (72GPH vs 408GPUH). 
\item Third, compared with the \textbf{LS-Evolution}~\cite{real2017large} on CIFAR-10 data, which achieves a test accuracy of 0.9180 with $17971$ GPUH, our \textbf{AG-Evolution} achieves a similar performance of $0.9052$ with much less time, i.e., 72GPUH. It is expected that by continuing our evolution process, we might achieve similar test accuracy to 0.9460 but with less amount of time. 
\item Finally,  \textbf{AG-Evolution} uses a much shorter time to find network structures achieving almost similar performance to \textbf{NA-Evolution} with a much longer time on SVHN, CIFAR-10 and CIFAR-100 datasets, which further verifies the benefit of the proposed aggressive selection. 
\end{itemize}%computational cost required by our genetic approach for obtaining better neural network structures is dramatically reduced compared to~\cite{real2017large} work spending $65,536$ GPU-Hours on CIFAR-10 and more on CIFAR-100 dataset, which is obviously not affordable to general users. 

\begin{figure}[h!]
 \centering
 \subfigure{\includegraphics[scale=0.14]{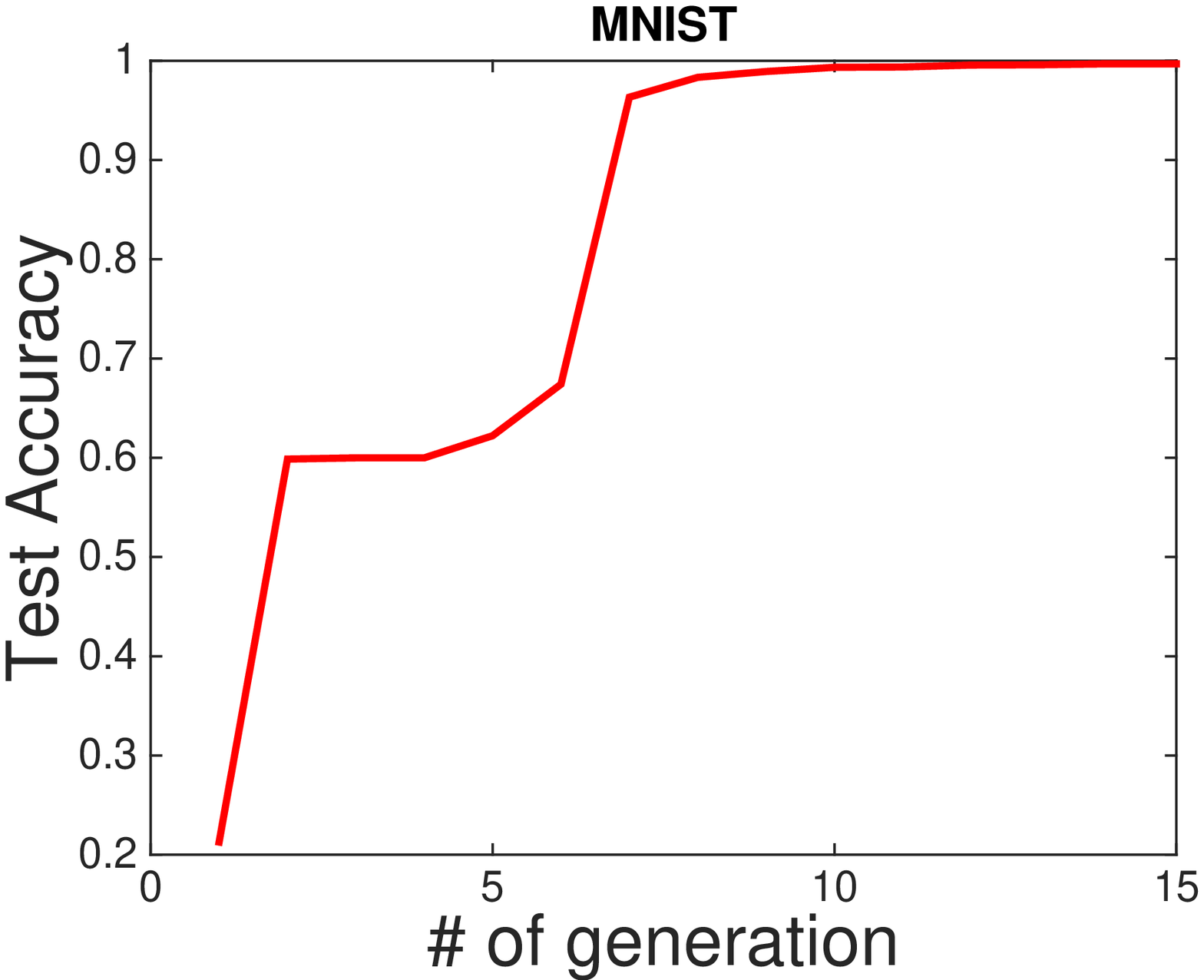}}%\hfill
\subfigure{\includegraphics[scale=0.14]{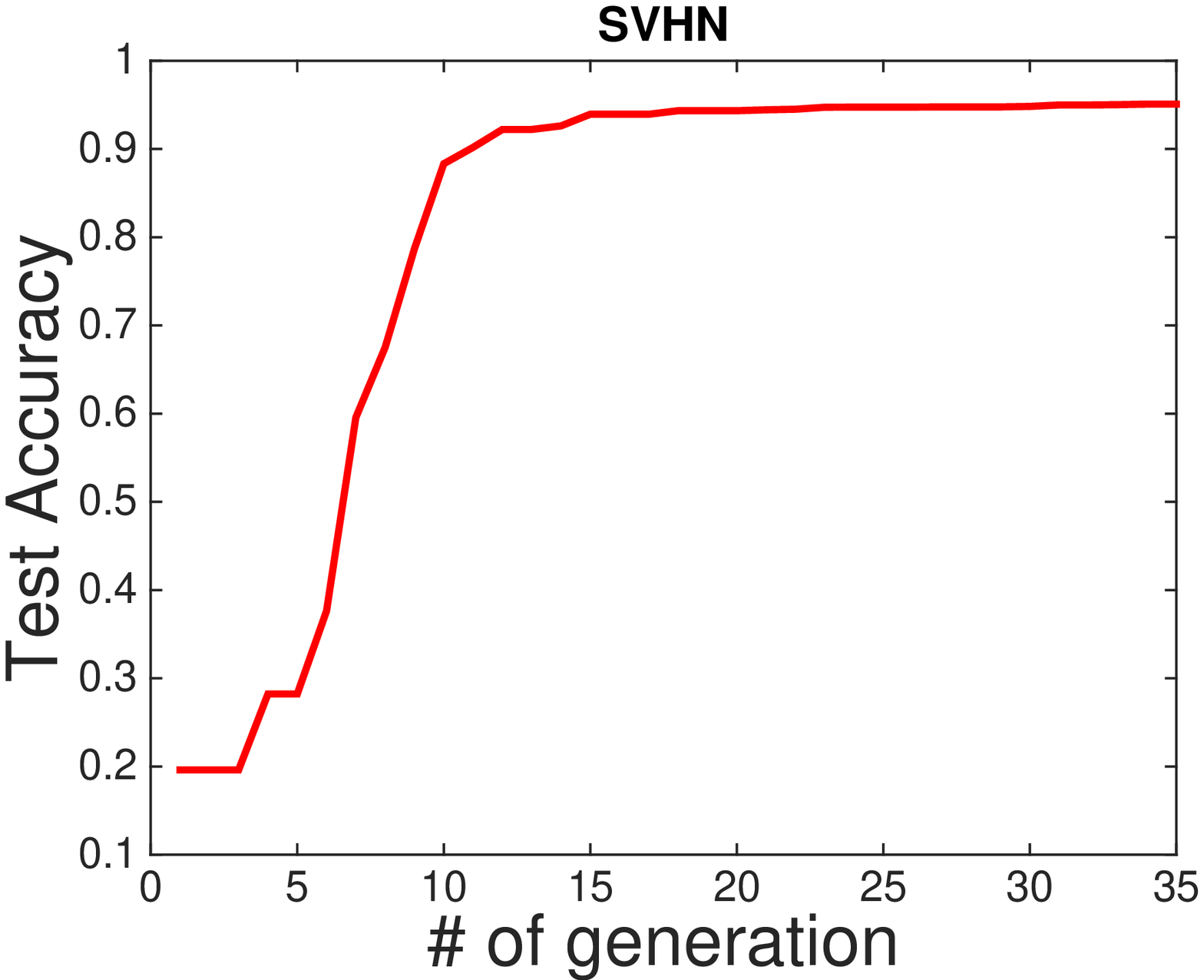}}
\subfigure{\includegraphics[scale=0.14]{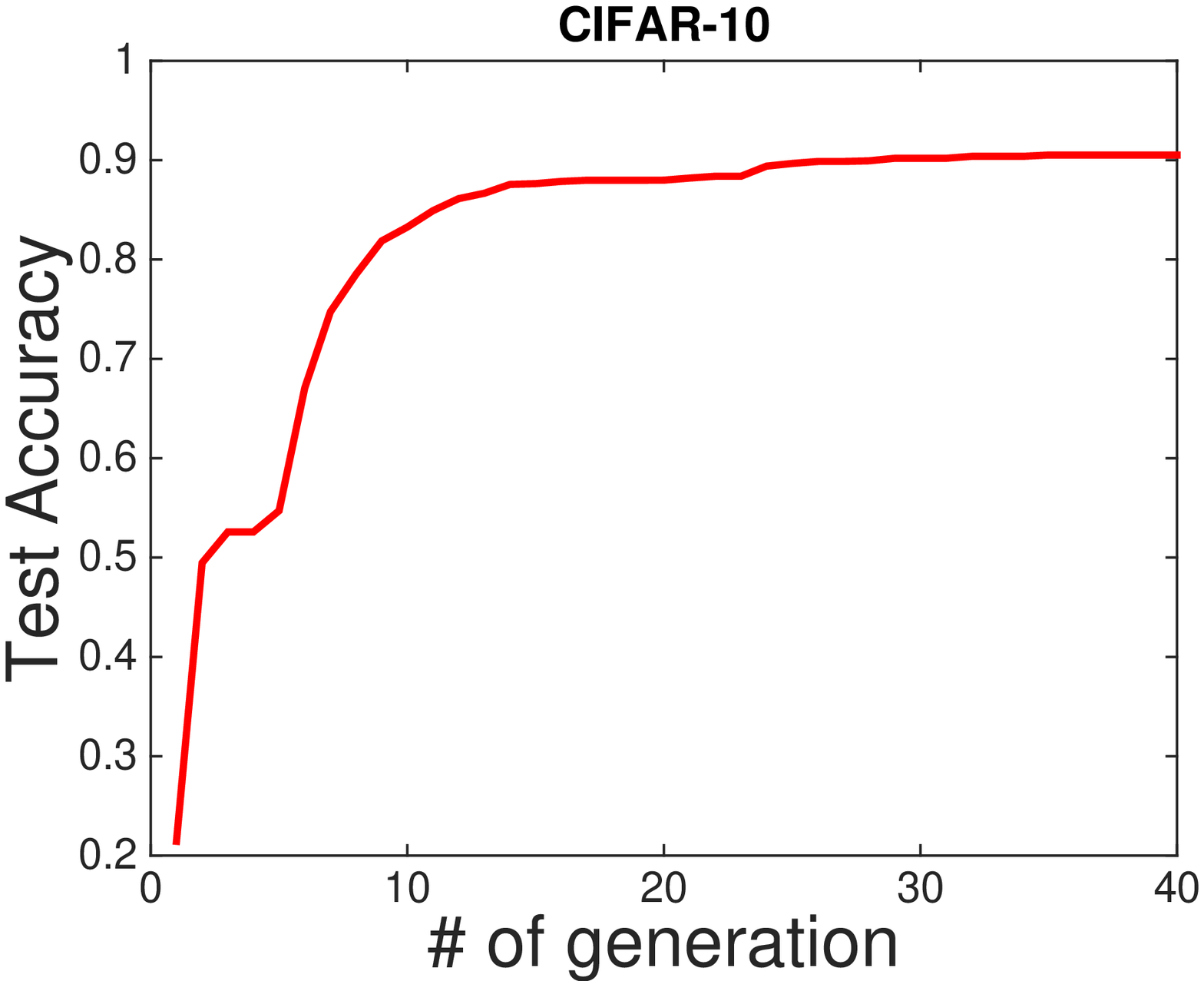}}
\subfigure{\includegraphics[scale=0.14]{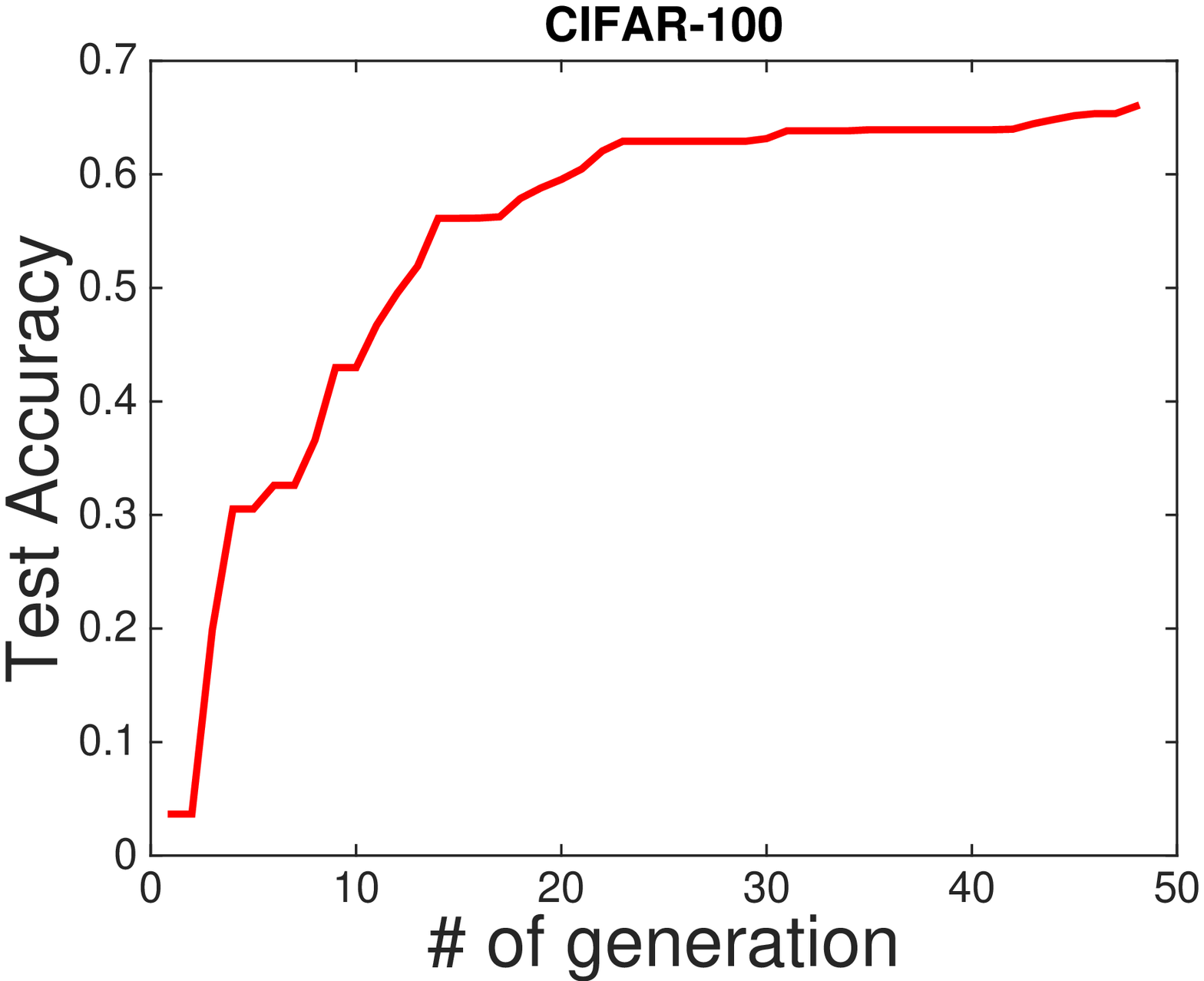}}
\caption{Test Accuracy vs Generation Number  on MNIST, SVHN, CIFAR-10, CIFAR-100 dataset}
 \label{fig:mnist-cifar10-svhn-cifar100-top1-acc}
\end{figure}

In Figure \ref{fig:mnist-cifar10-svhn-cifar100-top1-acc}, we plot the test accuracy of the best individual in each generation versus the number of generations on the four datasets for \textbf{AG-Evolution}, from which we could see the proposed genetic approach gradually improves the performance of neural network structures.  We also report the evolution of model size of individuals on CIFAR-10 dataset in the right of Figure~\ref{fig:cifar10-acc-topk}.

\begin{table}[t]
\begin{center}
\begin{tabular}{l|ll}
\hline
Approach     & \textbf{Test Acc} & \textbf{Comp Cost}  \\ 
\hline\hline
\textbf{SOTA}~\cite{wan2013regularization} & 0.9979     & --           \\ 
\textbf{Genetic-CNN}~\cite{xie2017genetic} & 0.9966     & 48 GPUH            \\ 
\textbf{EDEN}~\cite{dufourq2017eden} & 0.9840& --\\
\textbf{AG-Evolution} & 0.9969    &35 GPUH             \\ 
\hline
\end{tabular}
\end{center}
\caption{Comparison of test accuracy and computational cost on {MNIST} dataset. \textbf{NA-Evolution} is not run on this dataset due to that \textbf{AG-Evolution} almost achieves the same performance as the state-of-the-art.}
\label{mnist-quantitive-comp}
\end{table}

\begin{table}[h]
\begin{center}
\begin{tabular}{l|ll}
\hline
Approach     & \textbf{Test Acc} & \textbf{Comp Cost}  \\ 
\hline\hline
\textbf{SOTA}~\cite{lee2016generalizing} & 0.9831     & --           \\ 
\textbf{NA-Evolution} & 0.9620     &552 GPUH    \\       
\textbf{AG-Evolution} & 0.9541     &60 GPUH     \\ 
\hline
\end{tabular}
\end{center}
\caption{Comparison of test accuracy and computational cost on {SVHN} dataset.}
\label{svhn-quantitive-comp}
\end{table}

\begin{table}[h]
\begin{center}
\begin{tabular}{l|ll}
\hline
Approach     & \textbf{Test Acc} & \textbf{Comp Cost}  \\ 
\hline\hline
\textbf{SOTA}~\cite{huang2016densely} & 0.9654     & -            \\ 
\textbf{LS-Evolution}~\cite{real2017large} &0.9460 &$>$65,000  GPUH \\
\textbf{LS-Evolution}~\cite{real2017large} &0.9180 &$>$17,500  GPUH \\
\textbf{Genetic-CNN}~\cite{xie2017genetic} & 0.7706  &408 GPUH         \\ 
\textbf{EDEN}~\cite{dufourq2017eden} & 0.7450& --\\
\textbf{NA-Evolution} & 0.9037     &552 GPUH             \\ 
\textbf{AG-Evolution} & 0.9052     &72 GPUH             \\ 
\hline
\end{tabular}
\end{center}
\caption{Comparison of test accuracy and computational cost on {CIFAR-10} dataset.}
\label{cifar10-quantitive-comp}
\end{table}

\begin{table}[h]
\begin{center}
\begin{tabular}{l|ll}
\hline
Approach     & \textbf{Test Acc} & \textbf{Comp Cost}  \\ 
\hline\hline
\textbf{SOTA}~\cite{huang2016densely} & 0.8280     & -             \\ 
\textbf{LS-Evolution}~\cite{real2017large} & 0.7700 & $>$65,000 GPUH\\ 
\textbf{NA-Evolution} & 0.6560     &552 GPUH    \\ 
\textbf{AG-Evolution} & 0.6804     &184 GPUH    \\ 
\hline
\end{tabular}
\end{center}
\caption{Comparison of test accuracy and computational cost on {CIFAR-100} dataset.}
\label{cifar100-quantitive-comp}
\end{table}

%------------------------------------------------------------------------
Note that the benefit of this work for practitioners in deep learning and computer vision community is that they can utilize our work to automatically search the optimal neural network structures for their own task with acceptable computing cost.
%We note that \cite{} transferred the learned structure to large scale data set and verified the effectiveness of network structure. However, in \cite{} they manually added 4 convolutional and 2 pooling layers, adjusted the number of filters at three stages to 256, 512, 512, at the end added the fully connected layers with drop rate 0.5. This mannually tuning neural network to achieve better performance is different from our goal of automatically designing network structures without any human effort at all.

\subsection{Discovered Network Structures}
\label{sec::exp::evolve}
\begin{figure}[h!]
\begin{center}
    \includegraphics[scale=0.26]{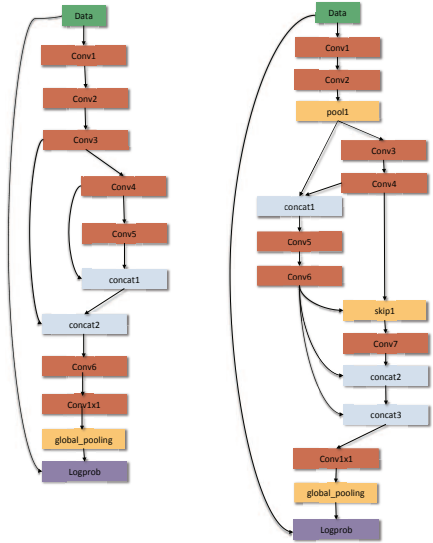}
\end{center}
\caption{Discovered neural network structures for CIFAR-10 and CIFAR-100 dataset.}
\label{fig:learned-neural-network-structure}
\end{figure}

Finally, we show two network structures found by the proposed \textbf{AG-Evolution} for CIFAR-10 and CIFAR-100 datasets in  Figure \ref{fig:learned-neural-network-structure}. %Interestingly, there is no fully connected layer in both networks. 
It is notable that both networks are multiple path networks with concatenation and skip connections similar to GoogleNet~\cite{googlenet15}  and ResNet~\cite{ResNet_cvpr16}.

%%%%% model size 
%\begin{figure}[h]
%\begin{center}
%   \includegraphics[scale=0.28]{figures/cifar10_model_size_acc_v0}
%\end{center}
%\caption{Model size, test accuracy vs generation number on CIFAR-10 dataset}
%\label{fig:cifar10_model_size_acc_V0}
%\end{figure}

%%%%% top 1 best worst bar 
%\begin{figure}[h]
%\begin{center}
   %\includegraphics[scale=0.32]{figures/example_learned_nn_cifar10_3}
%    \includegraphics[scale=0.32]{figures/top1_best_worst_bar.eps}
%\end{center}
%\caption{top1 best bar }
%\label{fig:top1_best_worst_bar}
%\end{figure}

\section{Conclusions}
In this paper, we have developed an aggressive genetic programing approach to optimize the structure of convolutional neural networks under limited computational resources without imposing strong restrictions on the search space. Our study shows that it is possible to achieve promising result using the proposed aggressive genetic programming approach in a reasonable amount of time. We expect the proposed strategies can be also useful for optimizing other types of neural networks (e.g., recurrent neural networks), which will be left as future work. %We will also explore whether better performance than the state-of-the-art on CIFAR-10, CIFAR-100 and large-scale ImageNet datasets can be achieved by our genetic programming approach by running more generations and  introducing more mutation operations. 

\bibliographystyle{splncs}
\bibliography{eccv2018submission}
\end{document}